\documentclass{vgtc}                          

\graphicspath{{figures/}{pictures/}{images/}{./}} 

\usepackage{times}                     

\usepackage{tabu}                      
\usepackage{booktabs}                  
\usepackage{lipsum}                    
\usepackage{mwe}                       

\usepackage{mathptmx}                  

\usepackage{algorithm}                    
\usepackage[noend]{algpseudocode}         
\usepackage{amsmath}                      
\usepackage{amssymb}                      

\usepackage{subfigure}

\onlineid{0}

\vgtccategory{Research}

\vgtcinsertpkg

\usepackage{hyperref}
\usepackage{graphicx}

\title{Extended Reality System for Robotic Learning from Human Demonstration}

\author{Isaac Ngui\thanks{e-mail: imngui2@illinois.edu}\\ %
        \scriptsize University of Illinois Urbana-Champaign %
\and Courtney McBeth\thanks{e-mail: cmcbeth2@illinois.edu}\\ %
     \scriptsize University of Illinois Urbana-Champaign %
\and Grace He\thanks{e-mail: gche3@illinois.edu}\\ %
    \scriptsize University of Illinois Urbana-Champaign %
\and André Corrêa Santos\thanks{e-mail: andrecs11@al.insper.edu.br}\\ %
    \scriptsize Insper %
\and Luciano Soares\thanks{e-mail: lpsoares@insper.edu.br} \\ %
    \scriptsize Insper %
\and Marco Morales\thanks{e-mail: moralesa@illinois.edu}\\ %
     \parbox{2in}{\scriptsize \centering ITAM \\ University of Illinois Urbana-Chmpaign} %
\and Nancy M. Amato\thanks{e-mail: namato@illinois.edu}\\ %
    \scriptsize University of Illinois Urbana-Champaign}

\teaser{
  \centering
  \includegraphics[width=\linewidth]{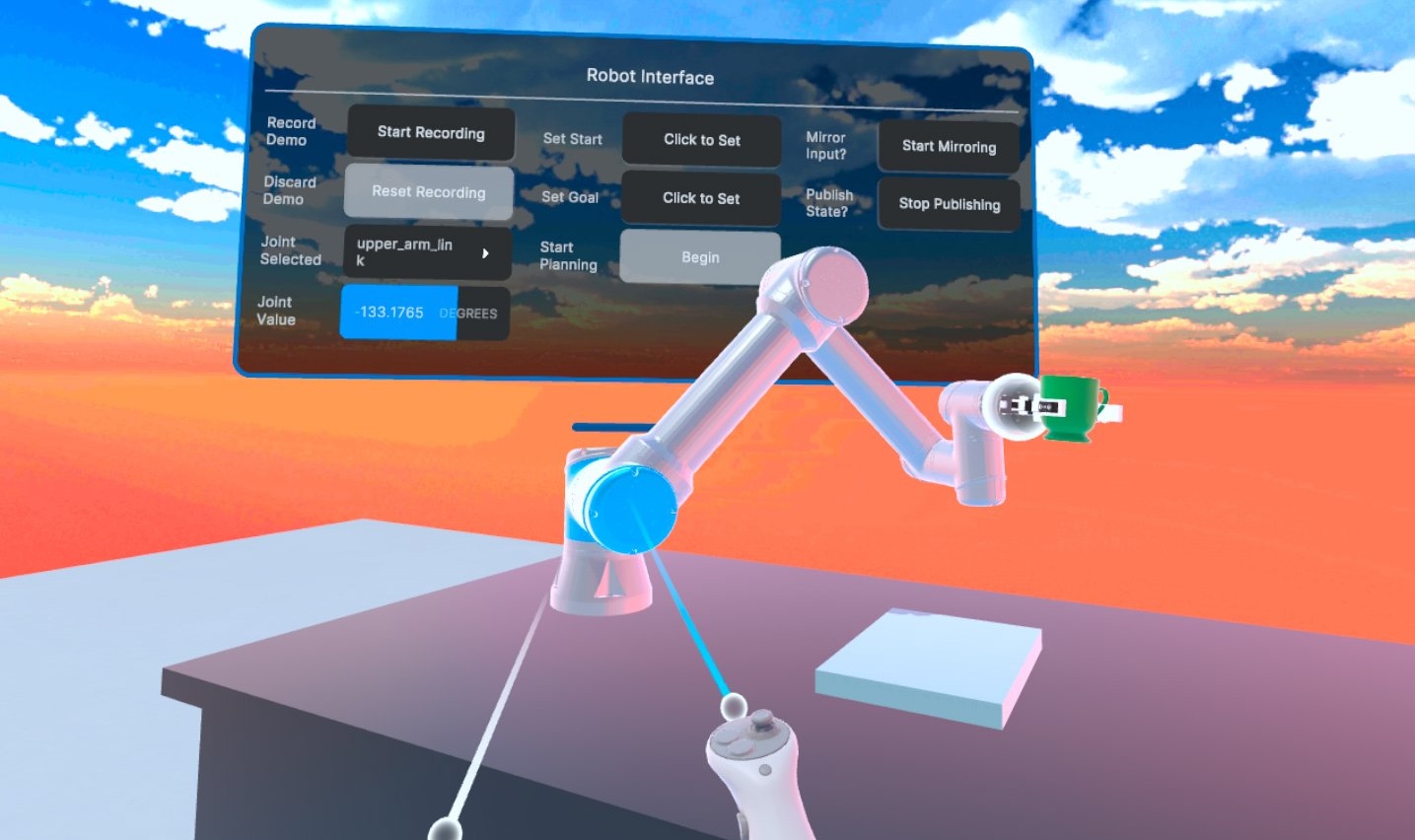}
  \caption{A human user interacting with a virtual UR5e robot to provide a trajectory demonstration as the robot carries a coffee mug over a table with a laptop on top.}
  \label{fig:teaser}
}

\abstract{
    Many real-world tasks are intuitive for a human to perform, but difficult to encode algorithmically when utilizing a robot to perform the tasks.
    In these scenarios, robotic systems can benefit from expert demonstrations, wherein human operators physically move the robot along trajectories, to learn how to perform each task.
    In many settings, it may be difficult or unsafe to use a physical robot to provide these demonstrations, for example, considering cooking tasks such as slicing with a knife.
    Extended reality provides a natural setting for demonstrating robotic trajectories while bypassing safety concerns and providing a broader range of interaction modalities.
    We propose the Robot Action Demonstration in Extended Reality (RADER) system, a generic extended reality interface for learning from demonstration.
    We additionally present its application to an existing state-of-the-art learning from demonstration approach and show comparable results between demonstrations given on a physical robot and those given using our extended reality system.

} 

\keywords{Robotics, extended reality, machine learning.}

\begin{document}



\maketitle

\section{Introduction} 

Robotic systems are becoming ubiquitous in many aspects of life ranging from industrial manufacturing settings to assistive technology in the home. Still, there are many tasks that humans can accomplish more efficiently and reliably than robots. Robots often struggle to accomplish fine-grained manipulation, perform complex motions, and adapt to human preferences when working in a shared space. In a domestic kitchen setting, for example, complex motions like mixing, scooping, and slicing~\cite{beltran2024sliceit} often pose a significant challenge.

This has led to the rise of \textit{Learning from Demonstration} (LfD) approaches, which aim to learn favorable robot behavior by training with human demonstrations. These demonstrations often consist of trajectories generated by the human via controller-based teleoperation or by directly dragging parts of the robot~\cite{lfd-survey}. They can be full executions of the desired task or corrections to the robot's existing behavior based on the human's preferences or safety concerns.

Many real-world tasks require the use of a manipulator robot, consisting of a chain of articulated bodies (Figure~\ref{fig:teaser}), along with a gripper to interact with objects in the world.
Controller-based demonstration systems are too simple to effectively capture the human user's desired motions. Many of these systems take joystick input to control only the \textit{end effector}, the tool mounted at the end of the arm, and compute the remaining joint angles using \textit{inverse kinematics}. This ignores the often nuanced preferences of the user who may want to restrict the robot's body from e.g. being near fragile objects. Other systems force the user to toggle the joint they are controlling using a button, which is time-consuming and does not result in fluid motions of the robot.

Demonstrations could also be provided in a fully simulated environment on a computer with a handheld controller or mouse and keyboard. This suffers from the same drawbacks as using a controller-based system with a physical robot, but additionally, results in the human user losing a level of understanding of what the robot's behavior will look like in practice due to the simulation's lack of immersion in reality.

Providing demonstrations by pushing or dragging a physical robot can often be difficult or unsafe for tasks involving heavy objects or sharp items, such as knives for slicing. 
The stiffness of robots' joints can inhibit those with physical impairments from being able to manually interact with them. 
Many commonly used robots only feature force-torque sensors on the end effector, meaning that feedback can only be provided by physically adjusting that link and not others, which again cannot capture nuanced preferences.
Additionally, some LfD methods~\cite{kalinowska2021ergodic} benefit from \textit{negative demonstrations} that place the robot in failure modes (e.g. in or near collision with itself or another object). These demonstrations cannot be safely provided with a physical robot.

\textit{Extended reality} (XR), which encapsulates \textit{virtual reality} (VR) and \textit{augmented reality} (AR), provides a modality in which these demonstrations can be safely and effectively provided. In VR, simulation environments can be created that replicate real-world applications like industrial manufacturing. AR, alternatively, provides the ability to visualize the behavior of a virtual robot in the physical environment before or concurrently with the deployment of a physical robot. Through interacting with a virtual robot, we bypass the limitations of existing systems to provide realistic demonstrations, enabling efficient and effective LfD.

We propose the \textit{Robot Action Demonstration in Extended Reality} (RADER) system, a novel interface to facilitate LfD by leveraging the capabilities of XR. Our contributions include:
\begin{itemize}
    \item RADER, a novel application of XR to robotics, allowing users to provide human demonstrations of robotic trajectories in VR or AR using natural controls without a lack of immersion.
    \item A modular interface, enabling users to easily switch between robots, virtual environments, and interface behaviors and visualizations.
    \item Experimental validation comparing the results of LfD using RADER to physical robot demonstrations showing the ability to achieve comparable results when using RADER-collected demonstrations as opposed to physical demonstrations.
    \item A publicly available, open-source implementation of our system, facilitating its use in future research projects and real-world applications.
\end{itemize}

\section{Preliminaries and Related Work}
In this section, we provide the background information necessary to understand our proposed system and discuss existing related work.

\subsection{Path and Trajectory Planning Preliminaries}
\begin{figure}[t]
    \centering
    \includegraphics[width=0.5\linewidth]{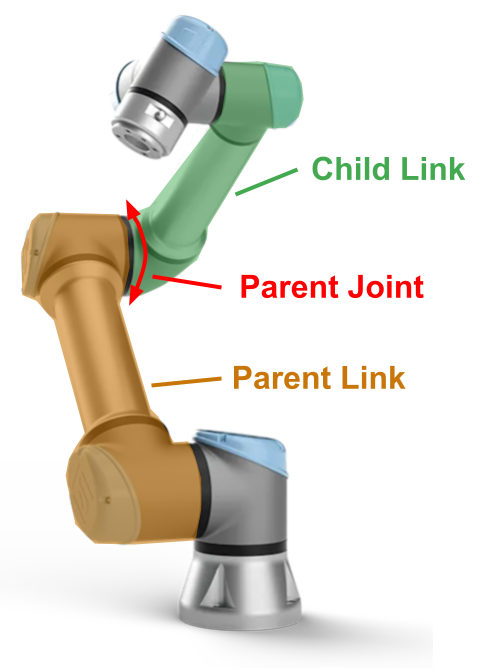}
    \caption{Diagram of manipulator arm components. A manipulator is composed of many links connected via joints. The orange link and the joint indicated by the red arrow are the parent link and joint respectively of the green link.}
    \label{fig:robot-parts}
\end{figure}

A \textit{configuration}, $q$, of a robot refers to a set of \textit{degree-of-freedom} values that completely describes each component's location in the environment. For a manipulator arm (Fig.~\ref{fig:robot-parts}), these degree-of-freedom values are the robot's joint angles. Given a \textit{query} consisting of a start configuration $q_s$ and a goal configuration $q_g$, the aim of \textit{path planning} is to find a sequence of intermediate waypoint configurations that make up a valid path $\pi = \{q_0, q_1, ..., q_{n-1} | q_0 = q_s, q_{n-1} = q_g\}$ from $q_s$ to $q_g$.
\textit{Trajectory planning} additionally ascribes a timing profile to $\pi$, resulting in a trajectory $T = \{(q_0, t_0), (q_1, t_1), ..., (q_{n-1}, t_{n-1}) | q_0 = q_s, q_{n-1} = q_g, t_0 = 0, t_{n-1} = 1 \}$ consisting of a sequence of configurations as well as times to reach them, generally normalized between 0 and 1. From the trajectory $T$, the desired robot controls can be derived.

\subsection{Learning from Demonstration}
There are many different approaches to using human demonstrations in learning-based trajectory planners. 
\textit{Behavioral cloning}~\cite{bain1999cloning} (BC) aims to learn a policy that replicates the behavior of the given human expert trajectories. BC has been applied to manipulation tasks including object pushing, scooping, and sweeping~\cite{florence2022implicitcloning}. Variations of BC have been introduced to account for sub-optimal non-expert demonstrations~\cite{sasaki2021noisy} and to include an initial self-supervised learning phase before conducting LfD, which can improve performance~\cite{faraz2018cloningobs}.

Another class of approaches to LfD is \textit{inverse reinforcement learning}~\cite{ng2000algorithms, arora2021irl, abbeel2004apprenticeship} (IRL), which aims to recover the underlying \textit{reward function} influencing the human demonstrations. The reward function is expressed in terms of \textit{features} describing the robot's observations and is a measure of the optimality of a trajectory subject to the human's implicit preferences. These preferences are often natural for a human to understand but difficult to programmatically specify, meaning they must be learned.

The \textit{Feature Expansive Reward Learning}~\cite{ferl} (FERL) method is a trajectory planning approach that recovers a reward function from demonstrated human corrections of robot behavior.
FERL generates and executes a trajectory that is optimal with respect to its current reward function over a state space consisting of information about the robot, such as joint angles, and objects in the environment.
The human may correct the movement of the robot during the execution by pushing it away from an undesirable or unsafe configuration or to apply a preference for how the robot should move.
Using the torques applied to each joint, FERL attempts to estimate its confidence in understanding what features the user was trying to express through the correction.
If the system is not confident in its understanding of the correction, the method assumes that a new feature is being expressed and the set of features is expanded.
FERL asks the human to provide a \textit{feature trace}, a trajectory demonstrating areas where the desired feature is strongly expressed and weakly expressed.
An updated reward function is computed using IRL and the trajectory is replanned given the updated reward function.
Providing these feature traces on a physical robot incurs the drawback of necessitating placing the robot into undesirable or unsafe configurations. We use FERL to experimentally show the advantages of providing demonstrations using RADER.

\subsection{Extended Reality in Robotics}
The fields of VR and robotics have long had a synergistic relationship~\cite{burdea1999synergy, freund1999bridginggap}. VR has wide applications in training operators of robotic systems ranging from surgical systems~\cite{bric2016surgery} and robotic prosthetics~\cite{frisoli2007exoskeleton} to industrial robotic systems~\cite{perez2019training} and educational settings~\cite{crespo2015learning, hurtado2021ur5vr}. In recent years, AR has become more prevalent in robotics applications~\cite{ar-survey}. AR has been used to communicate a robot's intent to a human user by projecting its expected path~\cite{andersen2016projecting, rosen2020communicating}. AR has also been widely used to enhance safety in human-robot interaction. Ostanin et al.~\cite{ostanin2020manipulator} use AR to visualize the reachable workspace of a robotic manipulator, informing human operators of potentially unsafe locations. Makhataeva et al.~\cite{makhataeva2019safetyaura} use AR to display a field around the robot showing areas a higher and lower estimated safety values.

Preliminary AR interfaces for human demonstration collection have also been explored. Leubbers et al.~\cite{luebbers2019augmented} present an interface that visualizes constraints that the robot is subject to while performing its task, enabling the human to comply with these constraints while providing their demonstration. This system, however, relies on a physical robot, limiting its potential use. Lotsaris et al.~\cite{ar-gripper} present an AR system that allows a user to position a virtual gripper in the environment. Here, the full robot trajectory is computed by a motion planner rather than provided by an expert human user. Similarly, George et al.~\cite{george2023openvr} propose a VR system for teleoperating a Franka Panda robot. Their approach only allows the user to manually set the pose of the end effector, limiting the interaction modalities accessible to the user. Iyer et al.~\cite{iyer2024open} propose a similar system that allows teleoperation based on hand gestures, but again does not allow the user to express their preferences through direct changes to the robot's joint angles.
Our system allows the human to provide complete robot trajectories using a virtual robot.

\section{Proposed System}
In this section, we describe RADER, our novel XR system to enable human users to provide realistic robotic trajectory demonstrations for LfD methods. RADER is intended to be robot-agnostic and generic with respect to trajectory planning methods to facilitate its use in a broad range of learning-based applications. Here, we first present RADER's system architecture and then describe our application of RADER to the FERL method as an example of its use.

\begin{figure}[t]
    \centering
    \includegraphics[width=1.0\linewidth]{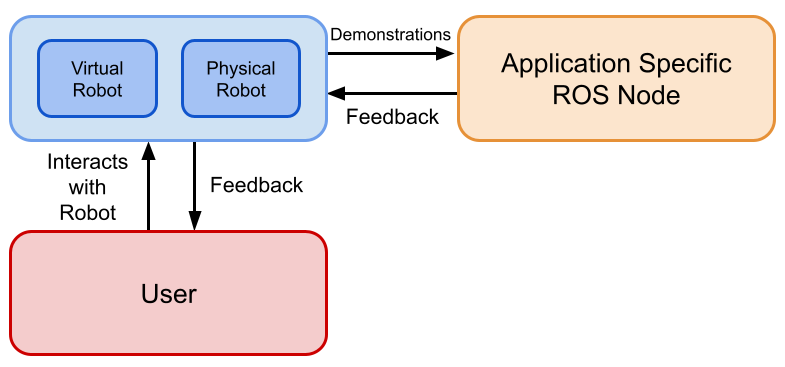}
    \caption{RADER system block diagram. As the human user interacts with the robot, collected trajectories are sent to application ROS nodes, which collect and process the data. Depending on the application, feedback may be provided by these nodes. Trajectories may be executed on a physical robot.}
    \label{fig:sys}
\end{figure}

\subsection{System Architecture}
RADER has three major components, the game engine that manages the XR interface, Robot Operating System (ROS) nodes that control the behavior of the virtual robot, and the human user that interacts with the virtual robot. An overview system diagram is shown in Fig.~\ref{fig:sys}.

\begin{figure}[t]
    \centering
    \includegraphics[width=0.75\linewidth]{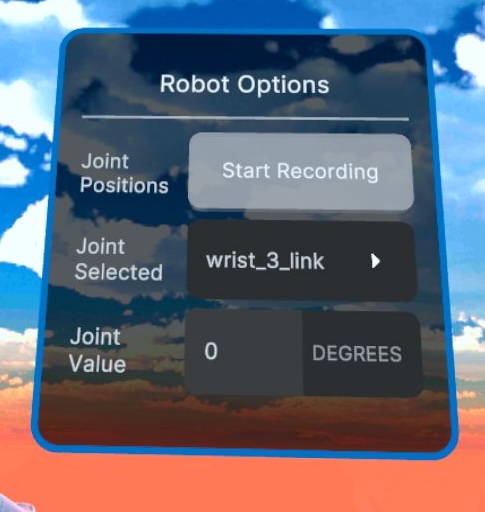}
    \caption{One of the default menus provided in RADER allows the user to select a joint from a drop-down menu and precisely set its angle value. This menu also provides the user the option to start and subsequently stop recording a demonstration.}
    \label{fig:menus}
\end{figure}

\begin{figure}[t]
    \centering
    \hfill
    \subfigure[Hover on Joint]{
        \includegraphics[width=0.22\textwidth]{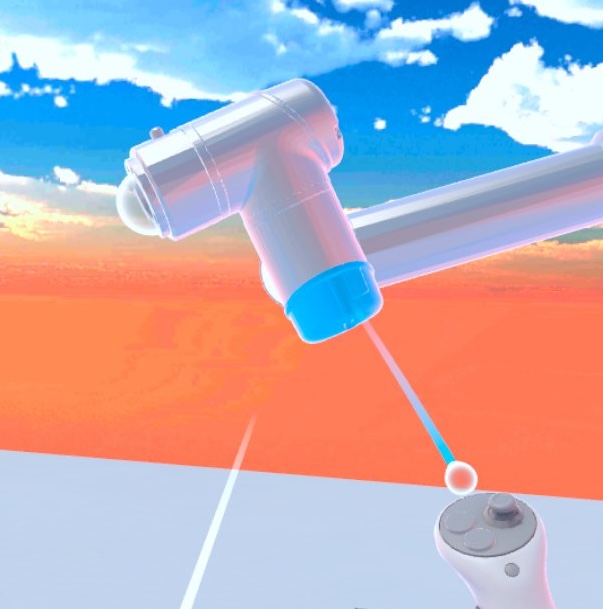}
    }
    \hfill
    \subfigure[Select Joint]{
        \includegraphics[width=0.22\textwidth]{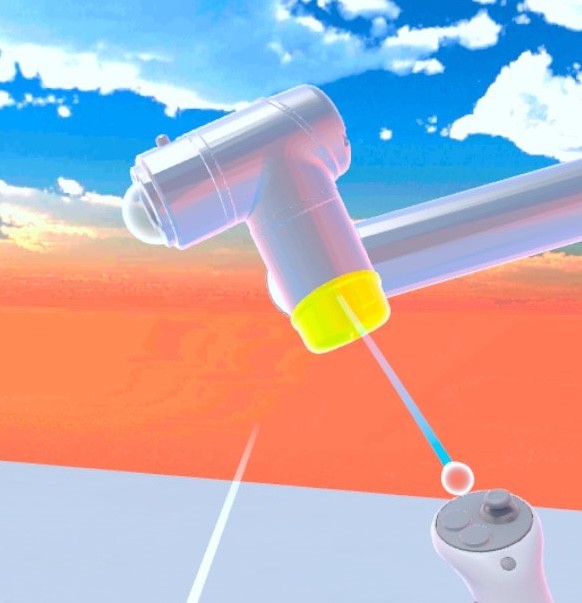}
    }
    \hfill
    \hfill
    \caption{The color of the mesh corresponding to a joint being hovered over or selected changes to inform the user of which joint they can interact with.}
    \label{fig:color-afford}
\end{figure}

\begin{figure}[t]
    \centering
    \includegraphics[width=1.0\linewidth]{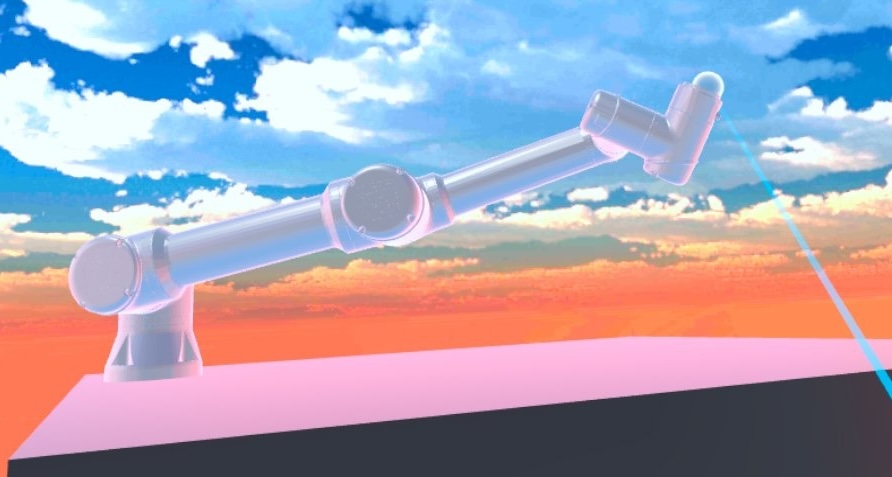}
    \caption{When the user moves the target sphere at the end of the end effector, inverse kinematics is used to find a robot configuration which places the end effector as close to the sphere as possible.}
    \label{fig:ik}
\end{figure}

\subsubsection{Game Engine}
We use Unity Game Engine\footnote{\url{https://unity.com}} to build our system due to its compatibility with a broad range of XR devices and wide availability of open-source packages. We release our Unity assets necessary for RADER as an open-source Unity package (see Sec.~\ref{sec:supplemental_materials}).

RADER supports robotic manipulator arms that are specified with unified robot description format (URDF) files, the standard format for modern commercial robots. The URDF file specifies the geometry for each of the robot's links and the articulation joint information. We use Unity's open-source URDF importer package\footnote{\url{https://github.com/Unity-Technologies/URDF-Importer}} and write scripts to support interacting with the imported robot. We also create a user interface consisting of several menu options (Fig.~\ref{fig:menus}) to modify the robot's joint angles, set the trajectory start and goal configurations, and start and stop recording demonstrations. Additional options can also be configured based on application-specific needs.

We support several different modes of interacting with the virtual robot to provide a demonstration. With either a VR controller or hand gestures, for supported devices, the user can grab the mesh corresponding to a link. By rotating the controller or their hand, the user can change the angle of the parent joint (Fig.~\ref{fig:robot-parts}) of the link. When hovered over or selected, a link is highlighted in a different color as an indication to the user (Fig.~\ref{fig:color-afford}). Alternatively, a user-selectable sphere (Fig.~\ref{fig:ik}) is placed at the end of the robot's end effector as an interactable target. The user can grab this target, again with either a controller or their hand, and move it around the environment. Inverse kinematics is used to continuously solve for robot configurations such that the end effector is placed as close as possible to the target, resulting in continuous motion of the robot. These different modalities provide different levels of control accessible to the user to express their preferences. More fine-grained control is available via directly modifying joint angles. This may be desired if the user wishes to prevent any links of the robot from reaching a specific area, for example. If the user's preferences are related to the end effector position, demonstrations can be more easily provided by dragging the inverse kinematics target sphere.

\subsubsection{Robot Operating System (ROS)}
ROS~\cite{macenski2022robot} is an open-source suite of robotics middleware software. ROS \textit{nodes} execute computations such as planning trajectories or updating learned models, send data to other nodes by \textit{publishing} to \textit{topics}, and receive data from other nodes by \textit{subscribing} to topics. Unity supports connecting to ROS nodes running on an external computer through a TCP connection\footnote{\url{https://github.com/Unity-Technologies/ROS-TCP-Connector}}. Using this connection, we send information about the virtual environment and robot to computation nodes and receive feedback. We use modularity and isolate the following functionalities to ensure they can be easily swapped out or built upon depending on the specific requirements of LfD method.

Trajectory planners require a representation of the environment the robot exists within. To ensure consistency between the virtual environment managed by the game engine and the external ROS nodes, point clouds are sampled from objects marked as obstacles in the virtual environment. These point clouds are then published to the external nodes. This allows the user to easily modify their environment by only interacting with the game engine without needing to update several representations.

Some LfD approaches (e.g.~\cite{ferl}) expect demonstrations in the form of corrections to the robot's motion while executing a computed trajectory. To accommodate this, we provide the ability to subscribe to a ROS topic that contains robot configurations published by external planner nodes. The virtual robot configuration is set based on these values which over time display a trajectory. The user can toggle this functionality on and off to provide their demonstration.

While collecting a demonstration, the robot's configuration is recorded to a list periodically in small time intervals and annotated with the current time. We also include approximated torque values applied to each of the joints at each time step using
$$\tau = I \times \alpha$$
where $I$ is the moment of inertia of the physical robot joint and $\alpha$ is the angular acceleration derived from the change in angle over time. More sophisticated methods could be used to calculate the torque if desired, however, we've found this to be sufficient for our experimental scenarios.
When the user indicates that their demonstration is completed, the compiled trajectory is published on a ROS topic to make it available to external ROS nodes running computation for the LfD method that collect and process the trajectory.

\subsubsection{Human User}
The role of the human user is to convey their implicit preferences to the LfD method by providing meaningful demonstrations. LfD methods often assume the human to be Boltzmann rational~\cite{ghosal2023effect} wherein they are more likely to choose actions that correspond to high utility. This behavior generally comes naturally to human users, although they are often limited in their ability to express it due to restrictions on the movements of physical robots. We aim to provide the user a broader range of action options in XR.

\begin{figure}[t]
    \centering
    \includegraphics[width=1.0\linewidth]{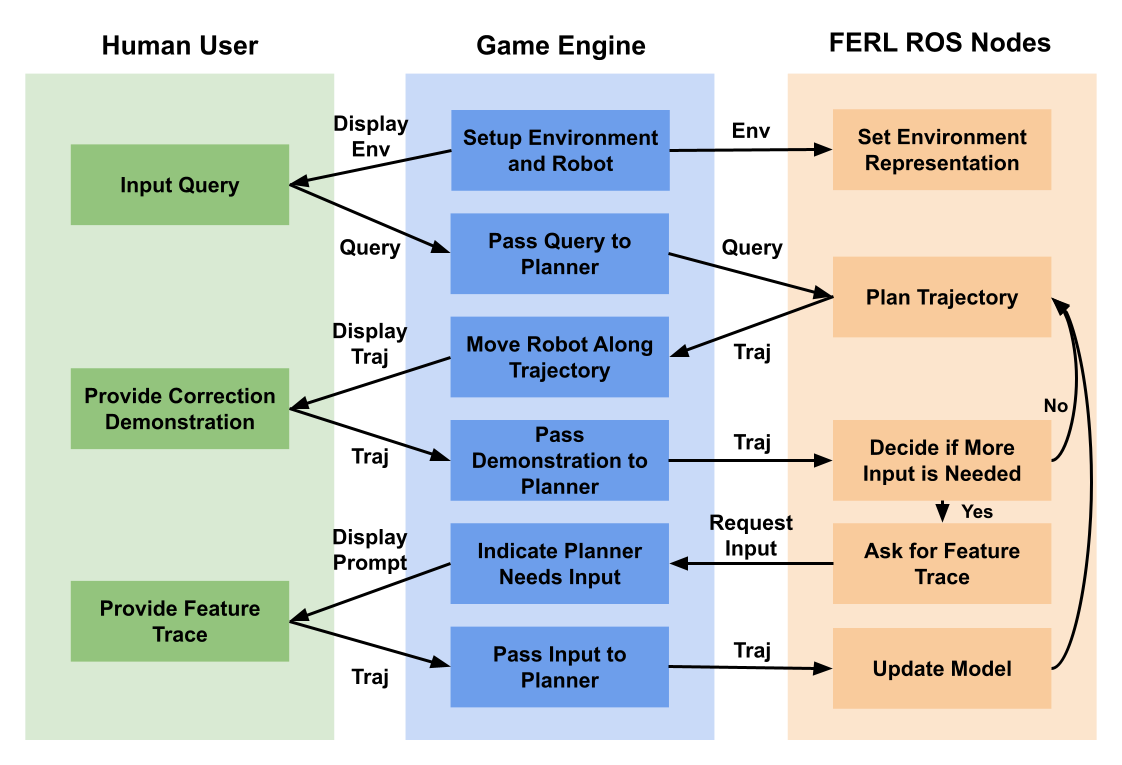}
    \caption{An event diagram showing the flow of events for each component of the system while executing FERL using RADER. \textit{Env} and \textit{Traj} are abbreviations of environment and trajectory, respectively.}
    \label{fig:ferl-flow}
\end{figure}

\subsection{Application to Feature Expansive Reward Learning}
To demonstrate its use, we apply RADER to LfD using FERL~\cite{ferl}. In this section, we describe how we built upon each component of the system to achieve this. Figure~\ref{fig:ferl-flow} shows the flow of events of the system.

\subsubsection{Game Engine}
FERL requires two types of demonstrations. The first type is demonstrations in the form of corrections to a trajectory the robot is executing. Given a correction, the algorithm estimates the confidence of its reward function with the current features to explain the demonstration. If the confidence is below a threshold, the algorithm then expands the set of features it considers and asks the user for demonstrations in the form of feature traces that indicate where areas of high and low reward are considering the new feature. To provide the ability for the algorithm to request feature traces, we create a script that subscribes to a ROS topic wherein requests to the user can be sent. If a message requesting a feature trace is received, a popup display appears prompting the user to input a feature trace demonstration.

\subsubsection{ROS Nodes}
An open-source codebase was provided by the authors of FERL~\cite{ferl} which we use as a foundation and adapt to interface with RADER. This codebase, as well as many other in the field, was built using ROS, so only minimal modifications to the software architecture are required. We add subscribers to read the user's inputted trajectory query and trajectory demonstrations and a publisher to request feature traces.

\subsubsection{Human User}
In FERL, the user is asked to provide corrections of the robot's movement and feature traces. In both of these cases, the user must provide meaningful data in order for the learned reward function to capture the user's preferences well. By providing demonstrations in XR, the user is able to place the robot in unsafe or undesirable configurations without damaging the robot or environment. This contributes to the user's ability to provide more expressive feature traces than would be possible on a physical robot.

\section{Experimental Validation}
To show the effectiveness of RADER in LfD applications, we compare the results of providing demonstrations using RADER against using a physical robot. We use a Meta Quest 3 headset as our XR device and a Universal Robots UR5e manipulator both in XR and as the physical robot for comparison.

\begin{figure}[t]
    \centering
    \includegraphics[width=0.75\linewidth]{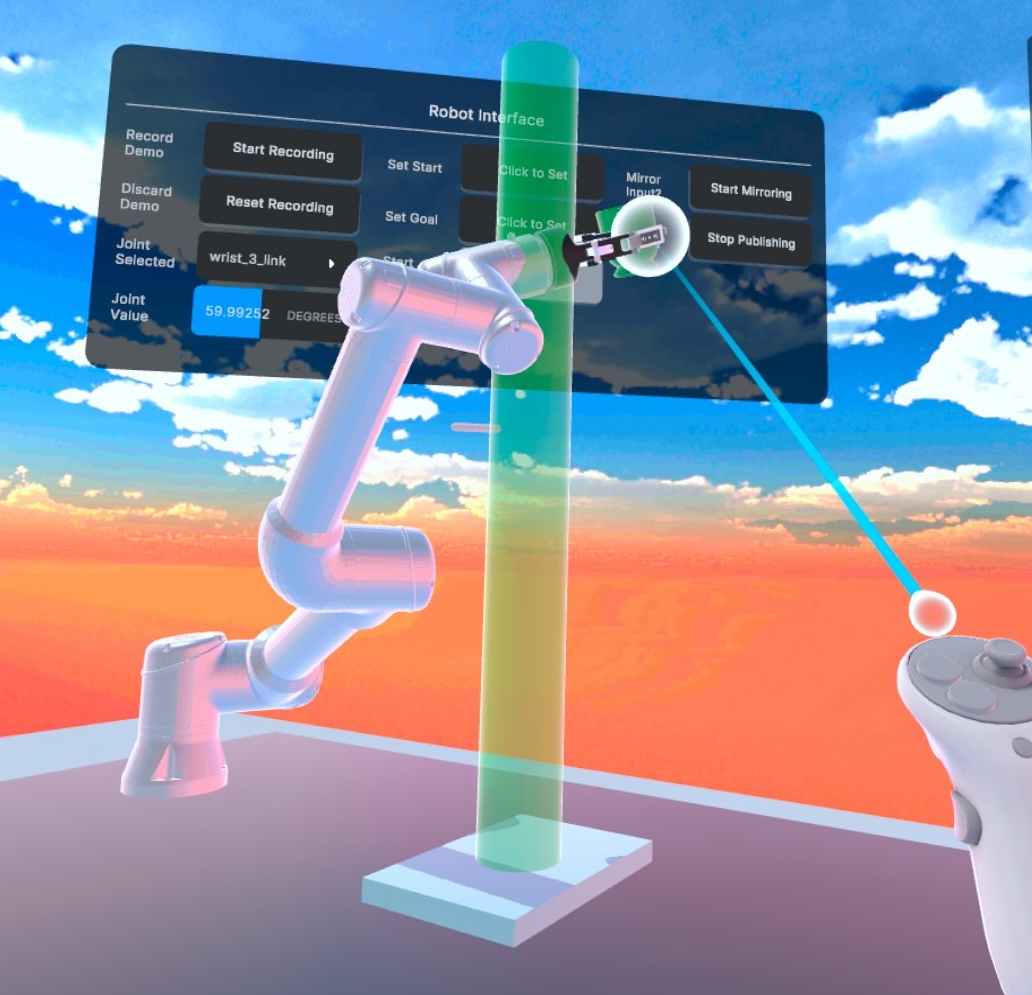}
    \caption{We place an indicator column above the laptop as a visual guide for the user to provide feature traces.}
    \label{fig:laptop-indicator}
\end{figure}

\subsection{Scenarios}
We use FERL~\cite{ferl} as the LfD method and replicate the experimental scenarios they present. The environment (Fig.~\ref{fig:teaser}) is set up such that the robot is placed on a table with a laptop on top of the table. The scenario is designed such that, when deployed, a human will stand to the side of the robot. We aim to learn features that convey that the robot should stay near the table (\textit{Table} feature), keep away from the laptop (\textit{Laptop} feature), and avoid being in close proximity to the human (\textit{Proxemics} feature).

\begin{figure}[t]
    \centering
    \includegraphics[width=1.0\linewidth]{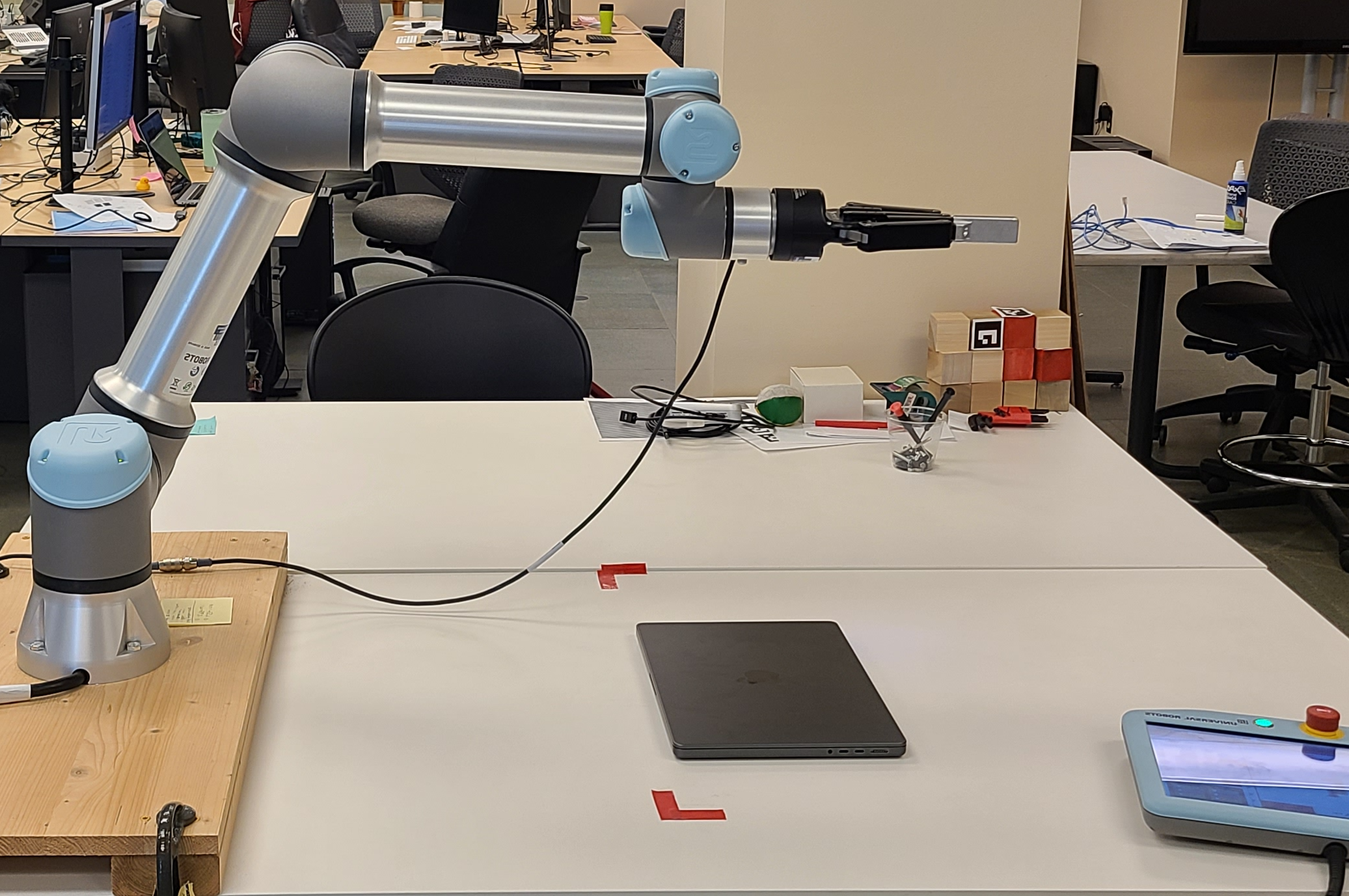}
    \caption{The physical UR5e robot setup to mirror the virtual robot in XR. The laptop is placed the same distance from the robot as in RADER.}
    \label{fig:phys-robot}
\end{figure}

\subsection{Physical Robot Setup}
The physical robot setup is designed to mimic the XR setup (Fig.~\ref{fig:phys-robot}).
The UR5e is placed at the end of a table in the same orientation as the robot in XR.
We utilize ROS2 Jazzy and the Universal Robotics ROS2 Driver\footnote{\url{https://github.com/UniversalRobots/Universal_Robots_ROS2_Driver}} package to handle low-level robot control.

To enable user interaction with the UR5e, we utilize force and torque sensor data from the robot to compute new end effector positions.
We use the MoveIt2 robotic manipulation platform\footnote{\url{https://moveit.picknik.ai}} to perform real-time servoing, end effector control operating at real-time speeds, to compute how the robot should react given the end effector force and torque reading.
The UR5e, like many other robotic manipulators, does not have support for joint-based force and torque control, the ability to interact with individual joints, and thus we use end effector-based control to mimic as much of this capability, however using RADER this is not a limitation.


\subsection{Results and Discussion}
\begin{figure}[t]
    \centering
    \includegraphics[width=1.0\linewidth]{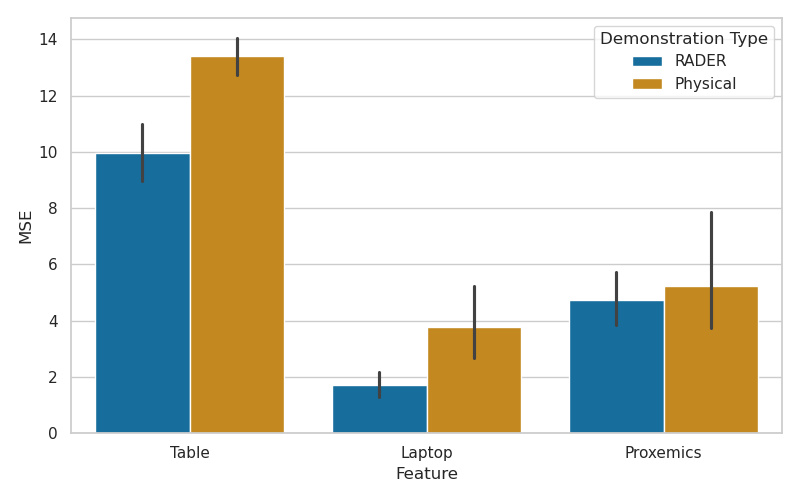}
    \caption{The mean squared error for the Table, Laptop, and Proxemics features for demonstrations provided using RADER and the physical robot. Error values are averaged over ten trials with ten feature traces sampled from the available set. The standard deviations are shown with error bars.}
    \label{fig:mse}
\end{figure}

We collect 20 feature traces for each of the Table, Laptop, and Proxemics features. We conduct ten trials per feature in which we randomly sample ten feature traces and learn the feature using FERL with a neural network consisting of 2 hidden layers with 64 neurons as performed in~\cite{ferl}. We show the mean squared error (MSE) for each learned feature in Figure~\ref{fig:mse}. In the sections below, we discuss these results.

\begin{figure}[t]
    \centering
    \includegraphics[width=1.0\linewidth]{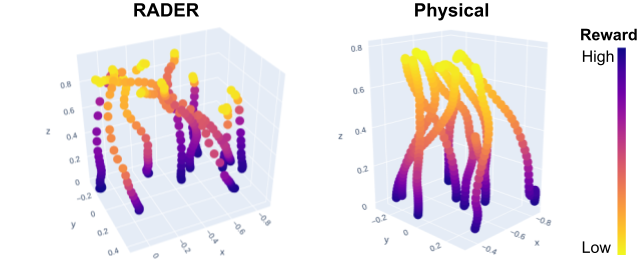}
    \caption{Ten example Table feature traces collected using RADER (left) and the physical robot (right). The feature traces are colored according to the corresponding learned reward function.}
    \label{fig:table-traces}
\end{figure}

\begin{figure}[t]
    \centering
    \includegraphics[width=1.0\linewidth]{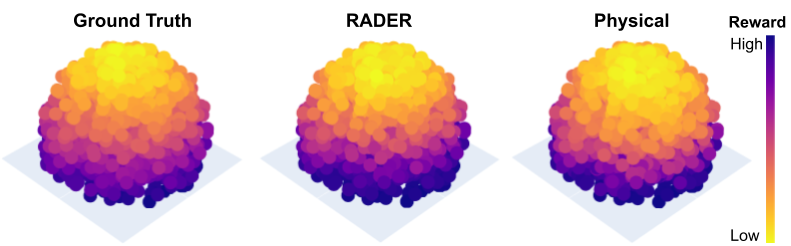}
    \caption{Plot of feature values over the reachable set of the robot. From left to right, the ground truth Table feature, the Table feature learned using RADER demonstrations, and the Table feature learned using physical robot demonstrations.}
    \label{fig:table-feature}
\end{figure}

\subsubsection{Table Feature}
The table feature captures the distance to the table, with a higher cost being associated with being further away from the table.
Example feature traces for the table feature are shown in Figure~\ref{fig:table-traces} and the learned feature is shown in Figure~\ref{fig:table-feature}. It can be seen from the learned feature plots and the MSE plots in Figure~\ref{fig:mse} that the RADER collected demonstrations have the ability to achieve comparable results to physical demonstrations for LfD. Both of the physical and RADER learned feature plots follow the ground truth gradient of feature values, achieving low reward when far away from the table and high reward when close to the table. This shows that the learned feature from demonstrations is rewarded for staying close to the table which matches the feature that was attempting to be captured.

\begin{figure}[t]
    \centering
    \includegraphics[width=1.0\linewidth]{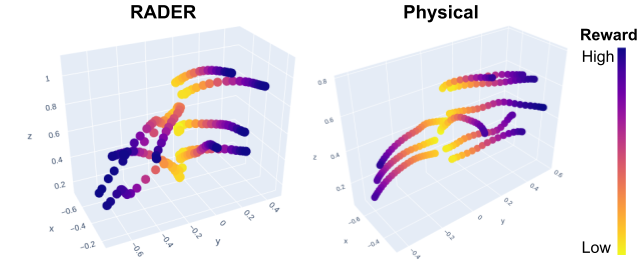}
    \caption{Ten example Laptop feature traces collected using RADER (left) and the physical robot (right). The feature traces are colored according to the corresponding learned reward function.}
    \label{fig:laptop-traces}
\end{figure}

\begin{figure}[t]
    \centering
    \includegraphics[width=1.0\linewidth]{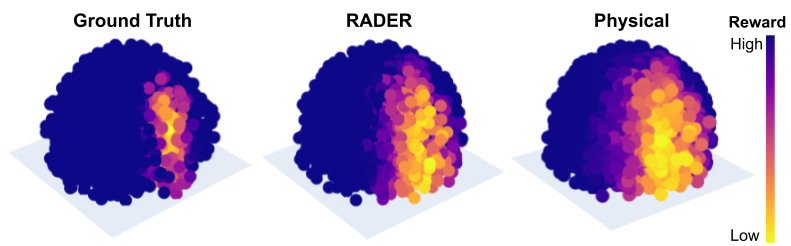}
    \caption{Plot of feature values over the reachable set of the robot. From left to right, the ground truth Laptop feature, the Laptop feature learned using RADER demonstrations, and the Laptop feature learned using physical robot demonstrations.}
    \label{fig:laptop-feature}
\end{figure}

\subsubsection{Laptop Feature}
As a visual guide, we place a translucent column above the laptop in XR (Fig.~\ref{fig:laptop-indicator}) to allow the human user to provide more precise feature traces, demonstrating another benefit of our system. Example feature traces are shown in Figure~\ref{fig:laptop-traces} and the feature learned using these feature traces is shown in Figure~\ref{fig:laptop-feature}. The indicator column allows the user to more accurately place the end effector at the beginning of each demonstration. When the end effector is positioned relatively high above the laptop, it can be difficult for the user to determine a precise start position in physical reality, contributing to a lower average MSE for the Laptop featured learned using RADER demonstrations.

\begin{figure}[t]
    \centering
    \includegraphics[width=0.9\linewidth]{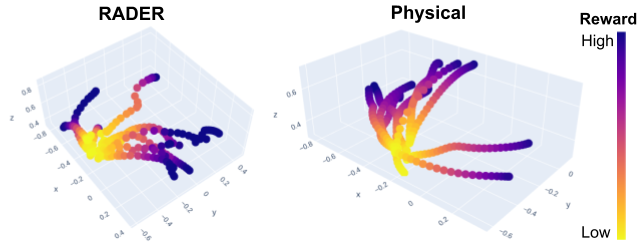}
    \caption{Ten example Proxemics feature traces collected using RADER (left) and the physical robot (right). The feature traces are colored according to the corresponding learned reward function.}
    \label{fig:proxemics-traces}
\end{figure}

\begin{figure}[t]
    \centering
    \includegraphics[width=1.0\linewidth]{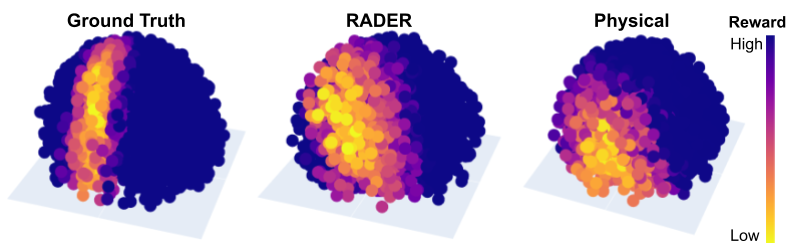}
    \caption{Plot of feature values over the reachable set of the robot. From left to right, the ground truth Proxemics feature, the Proxemics feature learned using RADER demonstrations, and the Proxemics feature learned using physical robot demonstrations.}
    \label{fig:proxemics-feature}
\end{figure}

\subsubsection{Human Proxemics Feature}
Example feature traces for the Proxemics feature are shown in Figure~\ref{fig:proxemics-traces} and the corresponding learned feature is shown in Figure~\ref{fig:proxemics-feature}. Here, the human operator is assumed to be positioned on the side of the table. The ground truth feature is computed by penalizing configurations that place the robot in front of the operator. The MSE values for the feature learned with RADER and physical demonstrations are very similar, showing the ability of RADER to generate valuable demonstrations. Both learned features tend to learn a broader area of low reward states. In practice, this can be explained by the human's preference for the robot to stay further away than is reflected in the ground truth.

\section{Conclusion}
We propose RADER, an XR system for demonstration collection for robotic LfD approaches. RADER is intended to be generic, accommodating a broad range of robots and LfD approaches. We present our foundational system architecture along with its application to a state-of-the-art LfD approach, FERL~\cite{ferl}. We show that providing demonstrations in XR allows the user to be more expressive with their demonstrations. Using FERL, a state-of-the-art LfD approach, we show the ability to generate valuable and realistic demonstrations in XR using RADER, comparing against those generated with a physical robot.

In the future, we intend to expand the functionality of RADER by adding support for mobile robots which will broaden the set of robotics applications we can consider. Although the demonstrations collected in this work were generated by the authors, going forward, we aim to conduct an extensive user study to determine whether those who are not experts in robotics can effectively and comfortably utilize our system. Additionally, we plan to expand RADER's user interface to accommodate the requirements of a broader set of demonstration types used by different LfD methods.

\section*{Supplemental Materials}
\label{sec:supplemental_materials}

Our code is made publicly available on GitHub under BSD 3-clause licenses. Specifically, the repository that contains the RADER Unity package can be found at \url{https://github.com/parasollab/RADER}. The VR project that contains the environment we used to run experiments can be found at \url{https://github.com/parasollab/VR_Robot}. The workspace containing our ROS nodes can be found at \url{https://github.com/parasollab/hri_ws}. Each of these repositories contains a README file with instructions to build and run the code contained within.




\section*{Figure Credits}
\label{sec:figure_credits}





\Cref{fig:robot-parts} image credit: Universal Robots \url{https://www.universal-robots.com/products/ur5-robot/} (modified).

\acknowledgments{
This work was supported in part by the Illinois-Insper Partnership at the University of Illinois and by Asociación Mexicana de Cultura AC.
}

\bibliographystyle{abbrv-doi}

\bibliography{refs}
\end{document}